\documentclass[11pt,a4paper]{article}
\usepackage[hyperref]{acl2021}
\usepackage{times}
\usepackage{latexsym}

\usepackage{amsmath}
\usepackage{multirow}
\usepackage{amssymb}
\usepackage{graphicx}

\usepackage{booktabs}
\usepackage{dcolumn}

\usepackage{microtype}

\aclfinalcopy 
\def\aclpaperid{***} 

\title{Cross-language Sentence Selection via \\ Data Augmentation and Rationale Training}

\author{Yanda Chen \\
  Columbia University \\
  \texttt{yc3384@columbia.edu} \\\And
  Chris Kedzie \\
  Columbia University \\
  \texttt{kedzie@cs.columbia.edu} \\\And
  Suraj Nair \\
  University of Maryland \\
  \texttt{srnair@umd.edu} \\ \And
  Petra Galuscakova \\
  University of Maryland \\
  \texttt{galuscakova@gmail.com} \\\And 
  Rui Zhang \\
  Yale University \\
  \texttt{r.zhang@yale.edu}
  }

\author{ 
  Yanda Chen\textsuperscript{1}, 
  Chris Kedzie\textsuperscript{1},
  Suraj Nair\textsuperscript{2},
  Petra Galu\v{s}\v{c}\'{a}kov\'{a}\textsuperscript{2},
  Rui Zhang\textsuperscript{3}, \\
  \bf {
  Douglas W. Oard\textsuperscript{2},
  Kathleen McKeown\textsuperscript{1}}\\
  \textsuperscript{1}Columbia University,
  \textsuperscript{2}University of Maryland,
  \textsuperscript{3}Penn State University\\ \\
  {\tt yc3384@columbia.edu, \{kedzie, kathy\}@cs.columbia.edu}\\
  {\tt \{srnair, petra, oard\}@umd.edu, rmz5227@psu.edu}
  }

\date{}

\newcommand{\qs}{E}
\newcommand{\ds}{S}
\newcommand{\dssize}{{|\ds|}}
\newcommand{\query}{q}
\newcommand{\rlabel}{r}
\newcommand{\dsprime}{\ds^\prime}
\newcommand{\qsprime}{\qs^\prime}
\newcommand{\dw}{w}
\newcommand{\qw}{\dw_q}
\newcommand{\relthr}{\lambda_{1}}
\newcommand{\data}{\mathcal{D}}
\newcommand{\vocab}{\mathcal{V}}
\newcommand{\datasize}{n}
\newcommand{\wemb}{W}
\newcommand{\embsize}{d}
\newcommand{\wembi}[1]{w_{#1}}
\newcommand{\dsi}{s}
\newcommand{\dsiprime}{\dsi^\prime}
\DeclareMathOperator*{\argmax}{arg\,max}
\newcommand{\amat}{A}
\newcommand{\mtrat}{\rho}
\newcommand{\modrat}{\alpha}
\newcommand{\logit}[2]{\wembi{#1}^\intercal\wembi{#2}}
\newcommand{\klloss}{\mathcal{L}_{rat}}
\newcommand{\kl}[2]{\operatorname{KL}(#1 \Vert #2)}
\newcommand{\relloss}{\mathcal{L}_{rel}}
\newcommand{\klweight}{\lambda_{2}}

\newcommand{\qsentprime}{E^\prime}
\newcommand{\qdata}{\mathcal{Q}}
\newcommand{\qvocab}{\mathcal{V}_{\qdata}}
\newcommand{\ddata}{\mathcal{S}}
\newcommand{\dvocab}{\mathcal{V}_{\ddata}}
\DeclareMathOperator*{\cossim}{cos-sim}

\newcommand{\word}{s}
\newcommand{\sent}{S}
\newcommand{\doc}{D}
\newcommand{\words}{{\word \in \sent}}
\newcommand{\sents}{{\sent \in \doc}}
\newcommand{\translation}{\hat{\word}}
\newcommand{\emb}[1]{w_{#1}}
\newcommand{\kernel}[2]{\emb{#1}^\intercal \emb{#2}}
\newcommand{\sizeop}[1]{{\lvert #1 \rvert}}
\newcommand{\Query}{Q}

\newcommand{\relevance}{\hat{r}}

\newtoggle{shownotes}
\toggletrue{shownotes}

\begin{document}
\maketitle
\begin{abstract}
This paper proposes an approach to cross-language sentence selection in a low-resource setting. It uses data augmentation and negative sampling techniques on noisy parallel sentence data to directly learn a cross-lingual embedding-based query relevance model. Results show that this approach performs as well as or better than multiple state-of-the-art machine translation + monolingual retrieval systems trained on the same parallel data. Moreover, when a rationale training secondary objective is applied to encourage the model to match word alignment hints from a phrase-based statistical machine translation model, consistent improvements are seen across three language pairs (English-Somali, English-Swahili and English-Tagalog) over a variety of state-of-the-art baselines.
\end{abstract}
\section{Introduction}

Sentence-level query relevance prediction is important for downstream  tasks such as query-focused summarization and open-domain question answering; accurately pinpointing sentences containing information that is relevant to the query  is critical to generating a responsive summary/answer (e.g., \citet{aaai1611939, baumel2018query}).
 In this work, we focus on sentence-level query relevance prediction in a cross-lingual setting, where the query and sentence collection are in different languages and the sentence collection is drawn from a low-resource language. Our approach enables English speakers (e.g., journalists) to find relevant information expressed in local sources (e.g., local reaction to the pandemic and vaccines in Somalia). 
  
While we can use machine translation (MT) to translate either the query or each sentence into a common language, and then use a monolingual Information Retrieval (IR) system to find relevant sentences, work on Probabilistic Structured Queries (PSQ) \cite{darwish2003probabilistic} has shown that the performance of such MT+IR pipelines is hindered by errors in MT. As is well known, complete translation of the sentence collection is not necessary. Inspired by previous work~\cite{vulic2015monolingual}, we go a step further and propose a simple cross-lingual embedding-based model that avoids translation entirely and directly predicts the relevance of a query-sentence pair (where the query and sentence are in different languages).

For training, we treat a sentence as \emph{relevant} to a query if there exists a translation equivalent of the query in the sentence. 
Our definition of relevance is most similar to the lexical-based relevance used in \citet{10.5555/1557769.1557825} and \citet{baumel2018query} but our query and sentence are from different languages. 
We frame the task as a problem of finding sentences that are relevant to an input query, and thus, we need relevance judgments for query-sentence pairs.
Our focus, however, is on low-resource languages where we have no sentence-level relevance judgments with which to train our query-focused relevance model.
We thus leverage noisy parallel sentence collections previously collected from the web.
We use a simple data augmentation and negative sampling scheme to generate a labeled dataset of relevant and irrelevant pairs of queries and sentences from these noisy parallel corpora. With this synthetic
training
set 
in hand, we can learn a supervised cross-lingual embedding space.

While our approach is competitive with pipelines of MT-IR, it is still sensitive to noise in the parallel sentence data.
We can mitigate the negative effects of this noise if we first train a phrase-based statistical MT (SMT) model on the same parallel sentence corpus and use the extracted word alignments as additional supervision. With these alignment hints, we demonstrate consistent and  significant improvements over neural and statistical MT+IR \cite{niu-etal-2018-bi, koehn2007moses, heafield-2011-kenlm}, three strong cross-lingual embedding-based models (Bivec \cite{luong-etal-2015-bilingual}, SID-SGNS \cite{Levy_2017}, MUSE \cite{conneau2017word}), a probabilistic occurrence model \cite{xu2000cross}, and a multilingual pretrained model XLM-RoBERTa \cite{conneau2019unsupervised}.
We refer to this secondary training objective as \emph{rationale training}, inspired by previous work in text classification that supervises attention over rationales for classification decisions~\cite{jain2019attention}.

To summarize, our contributions are as follows. We \emph{(i)} propose a data augmentation and negative sampling scheme to create a synthetic
training 
set
of cross-lingual query-sentence pairs with binary relevance judgements, and \emph{(ii)} demonstrate the effectiveness of a Supervised Embedding-based Cross-Lingual Relevance (SECLR) model trained on this data for low-resource sentence selection tasks on text and speech. Additionally, \emph{(iii)} we propose a rationale training secondary objective to further improve SECLR performance, which we call SECLR-RT.
Finally, \emph{(iv)} we conduct training data ablation and hubness studies that show our method's applicability to even lower-resource settings and 
mitigation of
hubness issues \cite{dinu2014improving, 10.1145/1835449.1835482}.
These findings are validated by empirical results of experiments in a low-resource sentence selection task, with English queries over sentence collections of text and speech in Somali, Swahili, and Tagalog.
\section{Related Work}
\label{sec:relatedworks}

\paragraph{Query-focused Sentence Selection} 
Sentence-level query relevance prediction is important for various downstream NLP tasks such as query-focused summarization \cite{aaai1611939, baumel2018query, 10.1145/3077136.3080690} and  open-domain question answering \cite{chen2017reading, dhingra2017quasar, 8554458}. Such applications often depend on a sentence selection system to provide attention signals on which sentences to focus upon to generate a query-focused summary or answer a question.

\paragraph{Cross-language Sentence Selection}
A common approach to cross-language sentence selection is to use MT to first translate either the query or the sentence to the same language and then perform standard monolingual IR~\cite{nie2010cross}. The risk of this approach is that errors in translation cascade to the IR system. 

As an alternative to generating full translations, PSQ \cite{darwish2003probabilistic} uses word-alignments from SMT to obtain weighted query term counts in the passage collection. In other work, \citet{xu2000cross} use a 2-state hidden Markov model (HMM) to estimate the probability that a passage is relevant given the query. 

\paragraph{Cross-lingual Word Embeddings}
Cross-lingual embedding methods perform cross-lingual relevance prediction by representing query and passage terms of different languages in a shared semantic space~\cite{vulic2015monolingual, 10.1145/3331184.3331324, litschko2018unsupervised, joulin-etal-2018-loss}. Both supervised approaches trained on parallel sentence corpora \cite{Levy_2017, luong-etal-2015-bilingual} and unsupervised approaches with no parallel data \cite{conneau2017word, artetxe-etal-2018-robust} have been proposed to train cross-lingual word embeddings.

Our approach differs from previous cross-lingual word embedding methods in two aspects. First, the focus of previous work has mostly been on learning a distributional word representation where translation across languages is primarily shaped by syntactic or shallow semantic similarity; it has not been tuned specifically for cross-language sentence selection tasks, which is the focus of 
our work. 

Second, in contrast to previous supervised approaches that train embeddings directly on a parallel corpus or bilingual dictionary, our approach trains embeddings on an \emph{artificial labeled dataset} augmented from a parallel corpus and directly represents relevance across languages. Our data augmentation scheme to build a relevance model is inspired by ~\citet{boschee-etal-2019-saral}, but we achieve significant performance improvement by incorporating rationale information into the embedding training process and provide detailed comparisons of performance with other sentence selection approaches. 
 
\paragraph{Trained Rationale}
Previous research has shown that models trained on classification tasks sometimes do not use the correct \emph{rationale} when making predictions, where a rationale is a mechanism of the classification model that is expected to correspond to human intuitions about salient features for the decision function \cite{jain2019attention}. Research has also shown that incorporating human rationales to guide a model's attention distribution can potentially improve model performance on classification tasks \cite{bao-etal-2018-deriving}. Trained rationales have also been used in neural MT (NMT); incorporating alignments from SMT to guide NMT attention yields improvements in translation accuracy \cite{chen2016guided}.
\section{Methods}
We first describe our synthetic 
training set
generation process, which
converts a parallel sentence corpus for MT into cross-lingual query-sentence pairs with binary relevance judgements for training our SECLR model.
Following that, we detail our SECLR model
and finish with our method for rationale training with word alignments from SMT.

\subsection{Training Set Generation Algorithm}
\label{datagen}

\textbf{\em Relevant query/sentence generation.}
Assume we have a parallel corpus of bilingual sentence pairs equivalent in meaning. Let $(\qs, \ds)$ be one such sentence pair, where $\qs$ is in the query language (in our case, English) and $\ds$ is in the 
retrieval collection language (in our case, low-resource languages).
For every unigram $\query$ in $\qs$ that is not a stopword, we construct a positive relevant sample by viewing $\query$ as a query and $\ds$ as a relevant sentence. Because sentences $\qs$ and $\ds$ are (approximately) equivalent in meaning, we know that there likely exists a translation equivalent of $\query$ in the sentence $\ds$ and so we label the $(\query, \ds)$ pair as relevant (i.e. $\rlabel=1$). 

For example, one English-Somali sentence pair is $\qs$=``true president gaas attend meeting copenhagen'', $\ds$=``ma runbaa madaxweyne gaas baaqday shirka copenhegan'' (stopwords removed). By extracting unigrams from $\qs$ as queries, we generate the following positive examples: ($\query$=``true'', $\ds$, $\rlabel=1$), ($\query$=``president'', $\ds$, $\rlabel=1$), ($\query$=``gaas'', $\ds$, $\rlabel=1$), ..., ($\query$=``copenhagen'', $\ds$, $\rlabel=1$). 

We generate the positive half of the training set by repeating the above process for every sentence pair in the parallel corpus. We limit model training to unigram queries since higher order ngrams appear fewer times 
and treating them independently reduces the risk of over-fitting. However, our model processes multi-word queries during evaluation, as described  in \autoref{sec:seclrmodel}.

\textbf{\em Irrelevant query/sentence generation.}
Since learning with only positive examples is a challenging task, we opt to create negative examples, i.e. tuples $(\query, \ds, \rlabel=0)$, via negative sampling. For each positive sample $(q, \ds, \rlabel=1)$, we randomly select another sentence pair $(\qsentprime, \dsprime)$ from the parallel corpus. We then check whether $\dsprime$ is relevant to $\query$ or not. 
Note that both the query $\query$ and sentence $\qsprime$ are in the same language, so checking whether $\query$ or a synonym can be found in $\qsprime$ is a monolingual task. If we can verify that there is no direct match or synonym equivalent of $\query$ in $\qsprime$ then by transitivity it is unlikely there exists a translation equivalent in $\dsprime$, making the pair $(\query,\dsprime)$ a negative example.
To account for synonymy when we check for matches, we represent $\query$ and the words in $\qsprime$ with pretrained word embeddings. Let $\qw, \dw_{\query^\prime} \in \mathbb{R}^d$ be the embeddings associated with $\query$ and the words $\query^\prime \in \qsprime$. We judge the pair $(\query,\dsprime)$ to be irrelevant (i.e. $\rlabel=0$) if:
\begin{equation*}
    \max_{\query^\prime \in \qsentprime} \cossim(\qw, \dw_{\query^\prime}) \le \relthr
\end{equation*} 
where $\relthr$ is a parameter. We manually tuned the relevance threshold $\relthr$ on a small development set of query-sentence pairs randomly generated by the algorithm, and set $\relthr=0.4$ to achieve highest label accuracy on the development set. 
 If $(\query, \dsprime)$ is not relevant we add $(\query, \dsprime, \rlabel=0)$ to our 
 synthetic training set, otherwise we re-sample $(\qsentprime, \dsprime)$ until a negative sample is found. We generate one negative sample for each positive sample to create a balanced dataset. 

For example, if we want to generate a negative example for the positive example ($\query$=``meeting'', $\ds$=``ma runbaa madaxweyne gaas baaqday shirka copenhegan'', $\rlabel=1$), we randomly select another sentence pair ($\qsentprime$=``many candidates competing elections one hopes winner'', $\dsprime$=``musharraxiin tiro badan sidoo u tartamaysa doorashada wuxuuna mid kasta rajo qabaa guusha inay dhinaciisa ahaato'') from the parallel corpus. To check whether $\query$=``meeting'' is relevant to $\dsprime$, by transitivity it suffices to check whether $\query$=``meeting'' or a synonym is present in $\qsprime$, a simpler monolingual task.
If $\query$ is irrelevant to $\dsprime$, we add $(\query, \dsprime, \rlabel=0)$ as a negative example. 

\subsection{Cross-Lingual Relevance Model}
\label{sec:seclrmodel}

We propose SECLR, a model that directly makes relevance classification judgments for queries and sentences of different languages without MT as an intermediate step by learning a cross-lingual embedding space between the two languages. Not only should translation of equivalent words in either language map to similar regions in the embedding space, but dot products  between query and sentence words should be correlated with the probability of relevance. We assume the training set generation process (\autoref{datagen}) provides us with a corpus of $\datasize$ query-sentence pairs along with their corresponding relevance judgements, i.e. $\data = \{(\query_i, \ds_i, \rlabel_i)\}|_{i=1}^{\datasize}$. We  construct a bilingual vocabulary $\vocab = \qvocab \cup \dvocab$ and associate with it a matrix $\wemb \in \mathbb{R}^{\embsize \times |\vocab| }$ where $\wembi{x} = \wemb_{\cdot, x}$ is the 
word embedding associated with word $x \in \vocab$.

When the query is a unigram $\query$ (which is
true 
by design in our training data $\data$), we model the probability of relevance to a sentence $\ds$ as:
\begin{equation*}
    p(\rlabel=1|\query, \ds; \wemb) = \sigma \left(\max_{\dsi \in \ds } \wembi{\query}^\intercal\wembi{\dsi} \right)
\end{equation*} 
where $\sigma$ denotes the logistic sigmoid ($\sigma(x) = 1 / \left(1 + \exp(-x)\right)$).

In our evaluation setting, the query is very often a phrase $\Query = [\query_1, \ldots, \query_\sizeop{\Query}]$. In this case, we require \emph{all} query words to appear in a sentence in order for a sentence to be considered as relevant. Thus, we modify our relevance model to be:
\begin{equation*}
     p(\rlabel=1|Q, \ds; \wemb) = \sigma \left(
\min_{\query \in Q} \max_{\dsi \in \ds } \wembi{\query}^\intercal\wembi{\dsi} \right)
\end{equation*}
Our only model parameter is the embedding matrix $\wemb$ which is initialized with pretrained monolingual word embeddings and learned via
minimization of the cross entropy of the relevance classification task:
\begin{equation*}
    \relloss = - \log p(\rlabel|\query, \ds;\wemb)
\end{equation*}

\subsection{Guided Alignment with Rationale Training}
\label{sec:rationale}

We can improve SECLR by incorporating additional alignment information as a secondary training objective, yielding SECLR-RT. 
Our intuition is that after training, the word 
$\translation = \argmax_{\words} \kernel{\word}{\query}$ should correspond
to a translation of $\query$. However, it is possible that $\translation$
simply co-occurs frequently with the true translation in our 
parallel data but its association is coincidental or irrelevant outside
the training contexts. We use alignment information to correct for this.
We run two SMT word alignment models, GIZA++ \cite{och2003} and
Berkeley Aligner \cite{haghighi-etal-2009-better}, on the orginal parallel sentence corpus. The two resulting alignments are concatenated as in \citet{zbib2019neural} to estimate a unidirectional probabilistic word translation matrix $\amat \in [0,1]^{|\vocab_Q| \times |\vocab_S|}$, such that $\amat$ maps each word in the query language vocabulary to a list of document language words with different probabilities, i.e. $\amat_{\query, \dsi}$ is the probability of translating $\query$ to $\dsi$ and $\sum_{\dsi \in \vocab_S} \amat_{\query,\dsi} = 1$.

For each relevant training sample, i.e. $(\query, \ds, \rlabel=1)$, we create a \emph{rationale distribution} $\mtrat \in [0,1]^{|\ds|}$ which is essentially a re-normalization of possible query translations
found in $\ds$ and represents our intuitions
about which words $\dsi \in \ds$ that $q$ should be most similar to in embedding space, i.e. 
\begin{equation*}
    \mtrat_\dsi = \frac{\amat_{\query,\dsi}}{\sum_{\dsiprime \in \ds}\amat_{\query,\dsiprime}}.
\end{equation*} 
for $\dsi \in \ds$. We similarly create a distribution under our model, $\modrat \in [0,1]^\dssize$, where 
\begin{equation*}
    \modrat_\dsi =  \frac{\exp\left(\wembi{\query}^\intercal\wembi{\dsi} \right)}{\sum_{\dsiprime \in \ds} \exp\left( \logit{\query}{\dsiprime} \right)} 
\end{equation*} for $\dsi \in \ds$.
To encourage $\modrat$ to match $\mtrat$, we impose a Kullback–Leibler (KL) divergence penalty, denoted as:
\begin{equation*}
    \klloss = \kl{\mtrat}{\modrat}
\end{equation*} to our overall loss function. The total loss for a single positive sample then will be a weighted sum of the relevance classification objective and the KL divergence penalty, i.e. 
\begin{equation*}
    \mathcal{L}=\relloss + \klweight \klloss
\end{equation*} where $\klweight$ is a relative weight between the classification loss and rationale similarity  loss. 

Note that we do not consider rationale loss for the following three types of samples: negative samples, positive samples where the query word is not found in the translation matrix, and positive samples where none of the translations of the query in the matrix are present in the source sentence. 
\section{Experiments}
\subsection{Dataset Generation from Parallel Corpus}
The parallel sentence data for training our proposed method and all baselines
includes the parallel data provided in the BUILD collections of both the MATERIAL\footnote{\url{https://www.iarpa.gov/index.php/research-programs/material}} and LORELEI
\cite{christianson2018overview}
programs for three low resource languages: Somali (SO), Swahili (SW),
and Tagalog (TL) (each paired with English). Additionally, we 
include in our parallel corpus
publicly available resources from OPUS 
\cite{tiedemann2012parallel},
and lexicons mined from Panlex
\cite{kamholz2014panlex}
and Wiktionary.\footnote{\url{https://dumps.wikimedia.org/}} Statistics of 
these parallel corpora and augmented data
are shown in \autoref{tab:pc-stat}  and \autoref{tab:augmented_dataset_stats}, respectively.
Other preprocessing details are in \autoref{app:data-details}.

\begin{table}[t]
\centering

\begin{tabular}{lrrr}
\toprule
 & EN-SO & EN-SW & EN-TL \\ 
\midrule
\# sents.  & 69,818 & 251,928 & 232,166  \\ 

EN tkn. & 1,827,826 & 1,946,556 & 2,553,439  \\ 

LR tkn. & 1,804,428 & 1,848,184 & 2,682,076  \\ 
\bottomrule
\end{tabular}
\caption{Parallel corpus statistics; ``EN tkn." refers to number of English tokens in the parallel corpus; ``LR tkn." refers to number of low-resource tokens (Somali, Swahili, Tagalog) in the parallel corpus. }
\label{tab:pc-stat}
\end{table}

\begin{table}[t]
\centering
\begin{tabular}{cc}
\toprule
Lang. Pair & Augmented Dataset Size\\
\midrule
EN-SO & 1,649,484\\

EN-SW & 2,014,838\\
EN-TL & 2,417,448\\ 
\bottomrule
\end{tabular}%
\caption{Augmented dataset statistics; ``augmented dataset size" refers to total number of positive and negative query-sentence samples in the augmented dataset.}
\label{tab:augmented_dataset_stats}
\end{table}

\begin{table*}[t]
\centering
\begin{tabular}{lccccccccc}
\toprule 
 \multirow{2}{*}{Lang.} &
   \multicolumn{3}{c}{Analysis} & \multicolumn{3}{c}{Dev} &
   \multicolumn{3}{c}{Eval} \\
   \cmidrule(lr){2-4} \cmidrule(lr){5-7} \cmidrule(lr){8-10}
  & \#Q & \#T & \#S & \#Q & \#T & \#S & \#Q & \#T & \#S \\ 
  \midrule 
 Somali &  300 & 338 & 142 & 300 & 482 & 213 & 1300 & 10717 & 4642\\
 Swahili & 300 & 316 & 155 & 300 & 449 & 217 & 1300 & 10435 & 4310\\
 Tagalog & 300 & 291 & 171 & 300 & 460 & 244 & / & / & /\\
   \bottomrule
\end{tabular}
\caption{MATERIAL dataset statistics: ``\#Q" refers to the number of queries; ``\#T" refers to the number of text documents; ``\#S" refers to the number of speech documents. There is no Tagalog Eval dataset.}
\label{tab:mat-stat}
\end{table*}

\begin{table*}[t]
\small
\begin{center}
\begin{tabular}{ l c c c c c  c  c c  c c  c  c }
\toprule
& \multicolumn{6}{c}{Somali} & \multicolumn{6}{c}{Swahili}\\ \cmidrule(lr){2-7}\cmidrule(lr){8-13} 
 & \multicolumn{2}{c}{Analysis} &  \multicolumn{2}{c}{Dev} & \multicolumn{2}{c}{Eval} & \multicolumn{2}{c}{Analysis} &  \multicolumn{2}{c}{Dev} & \multicolumn{2}{c}{Eval}  \\  
 \cmidrule(lr){2-3} \cmidrule(lr){4-5} \cmidrule(lr){6-7} \cmidrule(lr){8-9} \cmidrule(lr){10-11} \cmidrule(lr){12-13}
Method & T &  S & T & S & T & S & T &  S & T & S & T & S\\
\midrule
Bivec & 19.6 & 16.2 & 15.0 & 12.0 & ~4.2 & ~4.5 & 23.9 & 22.7 & 21.9 & 21.6 & 6.2 & 4.8 \\

SID-SGNS & 25.5 & 24.3 & 22.2 & 16.0 & 10.2 & ~9.1 & 38.8 & 36.3 & 33.7 & 30.3 & 16.2 & 13.6\\

MUSE & ~9.9 & ~9.9 & 10.3 & 16.5 & ~1.9 & ~2.0 & 27.8 & 24.5 & 27.3 & 28.8 & 9.5 & 8.1\\
\midrule
NMT+IR  & 18.8 & 12.5 & 21.1 & 13.4 & ~9.4 & ~8.4 & 23.7 & 24.9 & 26.8 & 26.7 & 15.3 & 11.4 \\
SMT+IR & 17.4 & 11.2 & 19.1 & 16.8 & ~9.1 & ~8.3 & 25.5 & 28.6 & 27.1 & 25.2 & 15.4 & 13.3\\
\midrule
PSQ & 27.0 & 16.6 & 25.0 & 20.7 & 11.1 & 8.6 & 39.0 & 36.6 & 38.0 & 38.6 & 20.4 & 13.8\\
\midrule
XLM-R & 13.9 & 11.0 & 10.7 & 12.4 & 2.3 & 2.9 & 23.3 & 29.0 & 20.0 & 29.7 & 6.2 & 7.5 \\
\midrule
SECLR & 27.8 & 24.4 & 23.0 & 17.4 & ~7.7 & ~7.4 & 43.8 & 37.9 & \textbf{40.3} & 38.1 & 16.0 & 13.1 \\
SECLR-RT & \textbf{35.4}\textsuperscript{$\dagger$} & \textbf{28.4} & \textbf{29.5} & \textbf{22.0} & \textbf{13.1}\textsuperscript{$\dagger$} & \textbf{11.2}\textsuperscript{$\dagger$} & \textbf{48.3}\textsuperscript{$\dagger$} & \textbf{48.1}\textsuperscript{$\dagger$} & 39.6 & \textbf{45.4} & \textbf{22.7}\textsuperscript{$\dagger$} & \textbf{17.7}\textsuperscript{$\dagger$}\\
\bottomrule
\end{tabular}
\caption{Document-level MAP scores for text (T) and speech (S) for Somali and Swahili. \textsuperscript{$\dagger$} indicates
significance at the $p=0.01$ level between SECLR-RT and the best 
baseline. 
}
\label{tab:mapmat_sosw}
\end{center}

\end{table*}

\begin{table}[t]
\begin{center}
\begin{tabular}{l c c  c  c }
\toprule
 & \multicolumn{2}{c}{Analysis} &  \multicolumn{2}{c}{Dev}\\  
  \cmidrule(lr){2-3} \cmidrule(lr){4-5}
Method & T &  S & T & S\\
\midrule
Bivec & 36.7 & 41.4 & 39.6 & 26.9 \\
SID-SGNS & 44.6 & 43.9 & 40.9 & 41.7 \\
MUSE & 27.4 & 26.5 & 26.0 & 16.5 \\
\midrule
NMT+IR & 37.7 & 42.3 & 32.6 & 37.5  \\
SMT+IR & 44.4 & 52.7 & 39.3 & 35.3 \\
\midrule
PSQ & 51.6 & 55.0 & 52.7 & 44.7 \\
\midrule
SECLR & 46.7 & 45.0 & 49.3 & 33.9 \\
SECLR-RT & \textbf{61.1} & \textbf{55.5} & \textbf{59.0} & \textbf{45.7}\\
\bottomrule
\end{tabular}
\caption{Document-level MAP scores for text (T) and speech (S) for Tagalog. 
}
\label{tab:mapmat_tl}
\end{center}
\end{table}

\subsection{Query Sets and Evaluation Sets}
\label{sec:qsetandeval}

We evaluate our sentence-selection model on English (EN) queries over three collections in  SO, SW, and TL recently made available as part of the IARPA MATERIAL program. 
In contrast to our training data which is synthetic, our evaluation datasets are human-annotated for relevance between real-world multi-domain queries and documents.
For each language there are three partitions (Analysis, Dev, and Eval),  with the former two being smaller collections intended for system development, and the latter being a larger evaluation corpus. In our main experiments we do not use Analysis or Dev for development and so we report results for all three (the ground truth relevance judgements for the TL Eval collection have not been released yet so we do not report Eval for TL). See \autoref{tab:mat-stat} for evaluation statistics. All queries are text. The speech documents are first transcribed with an ASR system \cite{Ragni2018}, and the 1-best ASR output is used in the sentence selection task.
Examples of the evaluation datasets are shown in \autoref{app:examples-of-eval}. 
We refer readers to \citet{rubino-2020-effect} for further details about MATERIAL test collections used in this work.

While our model and baselines work at the sentence-level, the MATERIAL relevance judgements are only at the document level. Following previous work on evaluation of passage retrieval, we aggregate our sentence-level relevance scores to obtain document-level scores~\cite{10.1145/278459.258561,Wade05passageretrieval,Fan_2018,10.1145/3269206.3271779,akkalyoncu-yilmaz-etal-2019-cross}. Given a document $\doc = [\sent_1, \ldots, \sent_\sizeop{\doc}]$, which is a sequence of sentences, and 
a query $\Query$, 
following \citet{10.1145/584792.584854} we assign a relevance score by:
\begin{equation*}
    \relevance = \max_\sents p(\rlabel=1|\Query, \sent; \wemb)
\end{equation*}

\subsection{Experiment Settings}
We initialize English word embeddings with word2vec \cite{mikolov2013efficient}, and initialize SO/SW/TL word embeddings with FastText \cite{grave2018learning}. For training we use a SparseAdam \cite{kingma2014adam} optimizer with learning rate
0.001. The hyperparameter $\lambda_2$ in \autoref{sec:rationale} is set to be 3 so that $\relloss$ and $\klweight \klloss$ are approximately on the same scale during training. More details on experiments are included in \autoref{app:supp}.

\subsection{Baselines}
\paragraph
{\em Cross-Lingual Word Embeddings.}
We compare our model with three other cross-lingual embedding methods, Bivec \cite{luong-etal-2015-bilingual}, MUSE \cite{conneau2017word}, and SID-SGNS \cite{Levy_2017}. Bivec and SID-SGNS are trained using the same parallel sentence corpus as the dataset generation algorithm used to train SECLR; thus, Bivec and SID-SGNS are trained on parallel sentences while SECLR is trained on query-sentence pairs derived from that corpus. We train MUSE with the bilingual dictionary from Wiktionary that is used in previous work \cite{Zhang_2019}. The SO-EN, SW-EN and TL-EN dictionaries have 7633, 5301, and 7088 words respectively. Given embeddings $\wemb^\prime$ from any of these methods, we compute sentence level relevance scores similarly to our model but use the cosine similarity: 
\begin{equation*}
    p(\rlabel=1|\Query,\sent;\wemb^\prime) =
    \min_{\query \in \Query} \max_\words \cossim(w^\prime_\word,w^\prime_\query)
\end{equation*}
since these models are optimized for this comparison function~\cite{luong-etal-2015-bilingual, conneau2017word, Levy_2017}. Document aggregation scoring is handled identically to our SECLR models (see \autoref{sec:qsetandeval}).

\paragraph
{\em MT+IR.} We also compare to a pipeline of NMT \cite{niu-etal-2018-bi} with monolingual IR and a pipeline of SMT~\footnote{We used Moses \cite{koehn2007moses} and KenLM for the language model \cite{heafield-2011-kenlm}.} with monolingual IR. Both MT systems are trained on the same parallel sentence data as our SECLR models. The 1-best output from each MT system is then scored with Indri \cite{strohmanICIA05} to obtain relevance scores. Details of NMT and SMT systems are included in \hyperref[app:mt-baseline]{Appendix}~\ref{app:mt-baseline}.

\paragraph
{\em PSQ.} 
To implement the PSQ model of \citet{darwish2003probabilistic}, we use the same alignment matrix as in rationale training (see \autoref{sec:rationale}) except that here we normalize the matrix such that $\forall \dsi \in \vocab_D, \sum_{\query \in \vocab_Q} \amat_{\query,\dsi} = 1$. Additionally, we embed the PSQ scores into a two-state hidden Markov model which smooths the raw PSQ scores with a background unigram language model \cite{xu2000cross}. The PSQ model scores each sentence and then aggregates the scores to document
level as in \autoref{sec:qsetandeval}.

\paragraph
{\em Multilingual XLM-RoBERTa.}
We compare our model to the cross-lingual model XLM-RoBERTa \cite{conneau2019unsupervised}, which in previous research has been shown to have better performance on low-resource languages than multilingual BERT \cite{devlin2018bert}. We use the Hugging Face implementation \cite{Wolf2019HuggingFacesTS} of XLM-RoBERTa (Base). We fine-tuned the model on the same augmented dataset of labeled query-sentence pairs
as the SECLR models, but we apply the XLM-RoBERTa tokenizer before feeding examples to the model. We fine-tuned the model for four epochs using an AdamW optimizer \cite{loshchilov2017decoupled} with learning rate $2\times 10^{-5}$. Since XLM-RoBERTa is pretrained on Somali and Swahili but not Tagalog, we only compare our models to XLM-RoBERTa on Somali and Swahili. 
\section{Results and Discussion}
\label{results}

We report Mean Average Precision (MAP) of our main experiment in \autoref{tab:mapmat_sosw} (SO \& SW) and \autoref{tab:mapmat_tl} (TL). Overall, we see that SECLR-RT consistently outperforms the other baselines in 15 out of 16 settings, and in the one case where it is not the best (SW Dev text), SECLR is the best. SECLR-RT is statistically significantly better than the best baseline on all Eval partitions.\footnote{We use a two-tailed paired t-test with Bonferroni correction for multiple comparisons at $p<0.01$ for all significance tests.} Since Analysis/Dev are relatively small, only three out of 12 Analysis/Dev settings are significant. The differences between SECLR and SECLR-RT can be quite large (e.g., as large as 70.4\% relative improvement on SO Eval text), suggesting that the rationale training provides a crucial learning signal to the model. 

Bivec and MUSE under-perform both of our model variants across all test conditions, suggesting that for the sentence selection task the relevance classification objective is more important than learning monolingual distributional signals. Curiously, SID-SGNS is quite competitive with SECLR, beating it on SO and SW Eval (both modalities) and TL Dev speech (five out of 16 test conditions) and is competitive with the other baselines. Again, the rationale training proves more effective as SID-SGNS never surpasses SECLR-RT.

While MT+IR is a competitive baseline, it is consistently outperformed by PSQ across all test conditions, suggesting that in low-resource settings it is not necessary to perform full translation to achieve good sentence selection performance. SMT, PSQ, and SECLR-RT all make use of the same word-alignment information but only SMT generates translations, adding additional evidence to this claim. 
PSQ and SECLR are close in performance on Analysis and Dev sets with SECLR eking out a slight advantage on seven of 12 Anaylsis/Dev set conditions.

On the larger Eval partitions, it becomes clearer that PSQ is superior to SECLR, suggesting that the relevance classification objective is not as informative as word alignment information. The relevance classification and trained rationale objectives capture slightly different information it seems; SECLR-RT, which uses both, out-performs PSQ across all 16 test conditions. 

\section{Training Data Ablation Study}

\begin{figure*}[ht]
\centering
\includegraphics[width=1.0\textwidth]{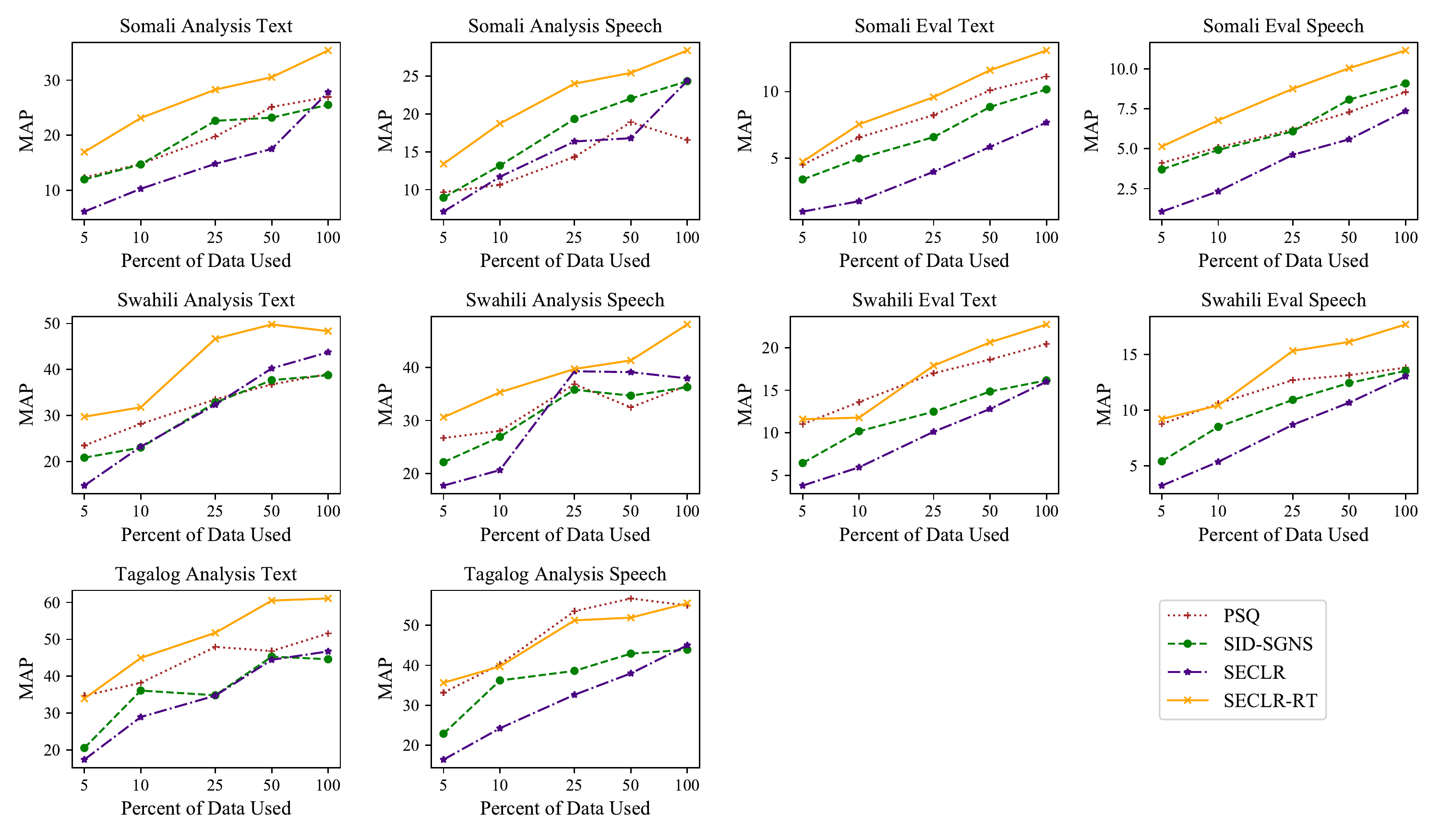}
\caption{Ablation study results of model performances as a function of sub-sampling percentages. Note that the x-coordinate uses the log scale for better illustration of low-resource cases. }
\label{ablation}
\end{figure*}

In \autoref{results}, we have shown that SECLR-RT consistently out-performs all baselines across all languages. Since this work targets cross-language sentence selection in a low-resource setting, we perform a training data ablation study to understand how training data size affects effectiveness.

We performed the ablation study for our two models SECLR and SECLR-RT, and the two strongest baseline methods PSQ and SID-SGNS. To simulate further 
the scenario of
data scarcity, we sub-sampled our parallel corpus uniformly at random for 5\%, 10\%, 25\%, 50\% of the sentence pairs of the original corpus. Each sentence pair in the parallel corpus is sampled with equal probability regardless of sentence length. For consistency, for each sample size, the same sampled parallel corpus is used across all models. The word alignment probability matrix used by PSQ and SECLR-RT is generated from the same sampled corpus. Since we tune the vocabulary size on the Dev set, for fair comparison we only report MAP scores on the Analysis and Eval sets. 

We plot MAP scores of the four models as a function of percentage of data sampled in \autoref{ablation}. Overall, we see that SECLR-RT consistently out-performs other baselines across all sample sizes in 9 out of 10 settings, and in the one case where it does not yield consistent improvement (Tagalog Analysis speech), SECLR-RT achieves comparable performance to PSQ. 

In the low-resource setting when the sample size is 5\% or 10\%, SECLR consistently under-performs other models, confirming our observation that SECLR is sensitive to noise and vulnerable to learning co-occurrences of word pairs that are in fact irrelevant. When the sample size is 5\% or 10\%, PSQ consistently achieves better performance than SID-SGNS and SECLR (although still under-performing SECLR-RT), indicating that alignment-based methods are more robust to noise and especially useful when data is extremely scarce. The fact that SECLR-RT consistently out-performs SECLR by a wide margin for small sample sizes indicates the 
necessity and
effectiveness of incorporating  alignment-based information into SECLR to improve the robustness of the model and learn more precise alignments. 
\section{Alleviating the Hubness Problem}

In this section, we show that by incorporating alignment information through rationale training, SECLR-RT significantly alleviates the hubness problem present in the trained cross-lingual embedding space produced by SECLR. 
Previous research on cross-lingual word embeddings has observed that a high-dimensional representation space with a similarity-based metric often induces a hub structure \cite{dinu2014improving}. Specifically, in a high-dimensional space (e.g., a cross-lingual word embedding space) defined with a pairwise similarity metric (e.g., cosine similarity), there exist a few vectors that are the nearest neighbors of many other vectors. Such vectors are referred to as ``hubs.'' The hub structure is problematic in IR since the hub vectors are often wrongly predicted as relevant and similar in meaning to queries that are in fact irrelevant \cite{10.1145/1835449.1835482}.

Let $\qvocab$ and $\dvocab$ be the embedding spaces for the query and sentence collection languages respectively. We define the size of the neighborhood of embeddings around $y \in \dvocab$ as \[N_k(y) = |\{x \in \qvocab | r_x(y) \leq k \}|\] where $r_x(y)$ is the rank of $y$ if we order $\dvocab$ by similarity to $x$ from highest to lowest, and $k$ is a positive integer. A large value of $N_k(y)$ indicates that $y$ is similar to many $x\in \qvocab$, and suggests that $y$ is a likely hub in embedding space.

Following \citet{10.1145/1835449.1835482}, we use $S_{N_{10}} = \mathbb{E}_{y \in \dvocab}[(N_{10}(y) - \mu)^3 / \sigma^3]$ to measure the skewness of the distribution of $N_{10}$, where $\mu$ and $\sigma$ refer to the mean and standard deviation of $N_{10}(y)$ respectively. Since cosine similarity is more frequently used as the similarity metric in hubness analysis, we re-train SECLR and SECLR-RT by replacing the dot product similarity metric with cosine similarity and still get performance comparable to \autoref{tab:mapmat_sosw} and \autoref{tab:mapmat_tl}.

We report $S_{N_{10}}$ scores for SECLR and SECLR-RT respectively in \autoref{tab:hub_deviation}. We see that SECLR-RT consistently has lower $S_{N_{10}}$ value compared to SECLR on all three languages, indicating that the extra alignment information incorporated with rationale training is helpful in reducing hubness.

\begin{table}[t]
\centering
\begin{tabular}{lccc}
\toprule
Model & Somali & Swahili & Tagalog \\ 
\midrule
SECLR & 29.36 & 54.98 & 43.29\\
SECLR-RT & \textbf{6.78} & \textbf{14.73} & \textbf{11.73}\\
\bottomrule
\end{tabular}
\caption{$S_{N_{10}}$ scores of SECLR and SECLR-RT respectively on Somali, Swahili and Tagalog. }
\label{tab:hub_deviation}
\end{table}

\section{Conclusion}
In this work, we presented a supervised cross-lingual embedding-based query relevance model, SECLR, for cross-language sentence selection and also applied a rationale training objective to further increase model performance. The resulting SECLR-RT model outperforms a range of baseline methods on a cross-language sentence selection task. Study of data ablation and hubness further indicate our model's efficacy in handling low-resource settings and reducing hub structures. 
In future work, we hope to apply our sentence-level query relevance approach to downstream NLP tasks such as query-focused summarization and open-domain question answering. 
\section*{Acknowledgements}
This research is based upon work supported in part
by the Office of the Director of National Intelligence (ODNI), Intelligence Advanced Research
Projects Activity (IARPA), via contract \#FA8650-
17-C-9117. The views and conclusions contained
herein are those of the authors and should not be interpreted as necessarily representing the official
policies, either expressed or implied, of ODNI,
IARPA, or the U.S. Government. The U.S. Government is authorized to reproduce and distribute
reprints for governmental purposes not withstanding any copyright annotation therein.

\bibliographystyle{acl_natbib}
\bibliography{acl2021}

\begin{thebibliography}{52}
\expandafter\ifx\csname natexlab\endcsname\relax\def\natexlab#1{#1}\fi

\bibitem[{Akkalyoncu~Yilmaz et~al.(2019)Akkalyoncu~Yilmaz, Yang, Zhang, and
  Lin}]{akkalyoncu-yilmaz-etal-2019-cross}
Zeynep Akkalyoncu~Yilmaz, Wei Yang, Haotian Zhang, and Jimmy Lin. 2019.
\newblock \href {https://doi.org/10.18653/v1/D19-1352} {Cross-{D}omain
  {M}odeling of {S}entence-{L}evel {E}vidence for {D}ocument {R}etrieval}.
\newblock In \emph{Proceedings of the 2019 Conference on Empirical Methods in
  Natural Language Processing and the 9th International Joint Conference on
  Natural Language Processing (EMNLP-IJCNLP)}, pages 3490--3496, Hong Kong,
  China. Association for Computational Linguistics.

\bibitem[{Artetxe et~al.(2018)Artetxe, Labaka, and
  Agirre}]{artetxe-etal-2018-robust}
Mikel Artetxe, Gorka Labaka, and Eneko Agirre. 2018.
\newblock \href {https://doi.org/10.18653/v1/P18-1073} {{A} {R}obust
  {S}elf-learning {M}ethod for {F}ully {U}nsupervised {C}ross-lingual
  {M}appings of {W}ord {E}mbeddings}.
\newblock In \emph{Proceedings of the 56th Annual Meeting of the Association
  for Computational Linguistics (Volume 1: Long Papers)}, pages 789--798,
  Melbourne, Australia. Association for Computational Linguistics.

\bibitem[{Bao et~al.(2018)Bao, Chang, Yu, and
  Barzilay}]{bao-etal-2018-deriving}
Yujia Bao, Shiyu Chang, Mo~Yu, and Regina Barzilay. 2018.
\newblock \href {https://doi.org/10.18653/v1/D18-1216} {Deriving {M}achine
  {A}ttention from {H}uman {R}ationales}.
\newblock In \emph{Proceedings of the 2018 Conference on Empirical Methods in
  Natural Language Processing}, pages 1903--1913, Brussels, Belgium.
  Association for Computational Linguistics.

\bibitem[{Baumel et~al.(2016)Baumel, Cohen, and Elhadad}]{aaai1611939}
Tal Baumel, Raphael Cohen, and Michael Elhadad. 2016.
\newblock \href
  {http://www.aaai.org/ocs/index.php/AAAI/AAAI16/paper/view/11939} {Topic
  {C}oncentration in {Q}uery {F}ocused {S}ummarization {D}atasets}.
\newblock In \emph{Proceedings of the Thirtieth {AAAI} Conference on Artificial
  Intelligence, February 12-17, 2016, Phoenix, Arizona, {USA}}, pages
  2573--2579. {AAAI} Press.

\bibitem[{Baumel et~al.(2018)Baumel, Eyal, and Elhadad}]{baumel2018query}
Tal Baumel, Matan Eyal, and Michael Elhadad. 2018.
\newblock \href {http://arxiv.org/abs/1801.07704} {Query {F}ocused
  {A}bstractive {S}ummarization: {I}ncorporating {Q}uery {R}elevance,
  {M}ulti-{D}ocument {C}overage, and {S}ummary {L}ength {C}onstraints into
  seq2seq {M}odels}.
\newblock \emph{CoRR}, abs/1801.07704.

\bibitem[{Boschee et~al.(2019)Boschee, Barry, Billa, Freedman, Gowda, Lignos,
  Palen-Michel, Pust, Khonglah, Madikeri, May, and
  Miller}]{boschee-etal-2019-saral}
Elizabeth Boschee, Joel Barry, Jayadev Billa, Marjorie Freedman, Thamme Gowda,
  Constantine Lignos, Chester Palen-Michel, Michael Pust, Banriskhem~Kayang
  Khonglah, Srikanth Madikeri, Jonathan May, and Scott Miller. 2019.
\newblock \href {https://doi.org/10.18653/v1/P19-3004} {{SARAL}: A
  {L}ow-{R}esource {C}ross-{L}ingual {D}omain-{F}ocused {I}nformation
  {R}etrieval {S}ystem for {E}ffective {R}apid {D}ocument {T}riage}.
\newblock In \emph{Proceedings of the 57th Annual Meeting of the Association
  for Computational Linguistics: System Demonstrations}, pages 19--24,
  Florence, Italy. Association for Computational Linguistics.

\bibitem[{Chen et~al.(2017)Chen, Fisch, Weston, and Bordes}]{chen2017reading}
Danqi Chen, Adam Fisch, Jason Weston, and Antoine Bordes. 2017.
\newblock \href {https://doi.org/10.18653/v1/P17-1171} {Reading {W}ikipedia to
  {A}nswer {O}pen-{D}omain {Q}uestions}.
\newblock In \emph{Proceedings of the 55th Annual Meeting of the Association
  for Computational Linguistics (Volume 1: Long Papers)}, pages 1870--1879,
  Vancouver, Canada. Association for Computational Linguistics.

\bibitem[{Chen et~al.(2016)Chen, Matusov, Khadivi, and Peter}]{chen2016guided}
Wenhu Chen, Evgeny Matusov, Shahram Khadivi, and Jan{-}Thorsten Peter. 2016.
\newblock \href {http://arxiv.org/abs/1607.01628} {Guided {A}lignment
  {T}raining for {T}opic-{A}ware {N}eural {M}achine {T}ranslation}.
\newblock \emph{CoRR}, abs/1607.01628.

\bibitem[{Christianson et~al.(2018)Christianson, Duncan, and
  Onyshkevych}]{christianson2018overview}
Caitlin Christianson, Jason Duncan, and Boyan~A. Onyshkevych. 2018.
\newblock \href {https://doi.org/10.1007/s10590-017-9212-4} {Overview of the
  {DARPA} {LORELEI} {P}rogram}.
\newblock \emph{Machine Translation}, 32(1-2):3--9.

\bibitem[{Conneau et~al.(2020)Conneau, Khandelwal, Goyal, Chaudhary, Wenzek,
  Guzm{\'a}n, Grave, Ott, Zettlemoyer, and Stoyanov}]{conneau2019unsupervised}
Alexis Conneau, Kartikay Khandelwal, Naman Goyal, Vishrav Chaudhary, Guillaume
  Wenzek, Francisco Guzm{\'a}n, Edouard Grave, Myle Ott, Luke Zettlemoyer, and
  Veselin Stoyanov. 2020.
\newblock \href {https://doi.org/10.18653/v1/2020.acl-main.747} {Unsupervised
  {C}ross-lingual {R}epresentation {L}earning at {S}cale}.
\newblock In \emph{Proceedings of the 58th Annual Meeting of the Association
  for Computational Linguistics}, pages 8440--8451, Online. Association for
  Computational Linguistics.

\bibitem[{Darwish and Oard(2003)}]{darwish2003probabilistic}
Kareem Darwish and Douglas~W. Oard. 2003.
\newblock \href {https://doi.org/10.1145/860435.860497} {Probabilistic
  {S}tructured {Q}uery {M}ethods}.
\newblock In \emph{Proceedings of the 26th Annual International ACM SIGIR
  Conference on Research and Development in Informaion Retrieval}, SIGIR '03,
  page 338–344, New York, NY, USA. Association for Computing Machinery.

\bibitem[{Devlin et~al.(2019)Devlin, Chang, Lee, and
  Toutanova}]{devlin2018bert}
Jacob Devlin, Ming-Wei Chang, Kenton Lee, and Kristina Toutanova. 2019.
\newblock \href {https://doi.org/10.18653/v1/N19-1423} {{BERT}: Pre-training of
  {D}eep {B}idirectional {T}ransformers for {L}anguage {U}nderstanding}.
\newblock In \emph{Proceedings of the 2019 Conference of the North {A}merican
  Chapter of the Association for Computational Linguistics: Human Language
  Technologies, Volume 1 (Long and Short Papers)}, pages 4171--4186,
  Minneapolis, Minnesota. Association for Computational Linguistics.

\bibitem[{Dhingra et~al.(2017)Dhingra, Mazaitis, and Cohen}]{dhingra2017quasar}
Bhuwan Dhingra, Kathryn Mazaitis, and William~W. Cohen. 2017.
\newblock \href {http://arxiv.org/abs/1707.03904} {{Q}uasar: {D}atasets for
  {Q}uestion {A}nswering by {S}earch and {R}eading}.
\newblock \emph{CoRR}, abs/1707.03904.

\bibitem[{Dinu and Baroni(2015)}]{dinu2014improving}
Georgiana Dinu and Marco Baroni. 2015.
\newblock \href {http://arxiv.org/abs/1412.6568} {{I}mproving {Z}ero-shot
  {L}earning by {M}itigating the {H}ubness {P}roblem}.
\newblock In \emph{3rd International Conference on Learning Representations,
  {ICLR} 2015, San Diego, CA, USA, May 7-9, 2015, Workshop Track Proceedings}.

\bibitem[{Fan et~al.(2018)Fan, Guo, Lan, Xu, Zhai, and Cheng}]{Fan_2018}
Yixing Fan, Jiafeng Guo, Yanyan Lan, Jun Xu, Chengxiang Zhai, and Xueqi Cheng.
  2018.
\newblock \href {https://doi.org/10.1145/3209978.3209980} {{M}odeling {D}iverse
  {R}elevance {P}atterns in {A}d-hoc {R}etrieval}.
\newblock In \emph{The 41st International {ACM} {SIGIR} Conference on Research
  {\&} Development in Information Retrieval, {SIGIR} 2018, Ann Arbor, MI, USA,
  July 08-12, 2018}, pages 375--384. {ACM}.

\bibitem[{Feigenblat et~al.(2017)Feigenblat, Roitman, Boni, and
  Konopnicki}]{10.1145/3077136.3080690}
Guy Feigenblat, Haggai Roitman, Odellia Boni, and David Konopnicki. 2017.
\newblock \href {https://doi.org/10.1145/3077136.3080690} {{U}nsupervised
  {Q}uery-{F}ocused {M}ulti-{D}ocument {S}ummarization {U}sing the {C}ross
  {E}ntropy {M}ethod}.
\newblock In \emph{Proceedings of the 40th International ACM SIGIR Conference
  on Research and Development in Information Retrieval}, SIGIR’17, page
  961–964, New York, NY, USA. Association for Computing Machinery.

\bibitem[{Grave et~al.(2018)Grave, Bojanowski, Gupta, Joulin, and
  Mikolov}]{grave2018learning}
Edouard Grave, Piotr Bojanowski, Prakhar Gupta, Armand Joulin, and Tom{\'{a}}s
  Mikolov. 2018.
\newblock \href
  {http://www.lrec-conf.org/proceedings/lrec2018/summaries/627.html}
  {{L}earning {W}ord {V}ectors for 157 {L}anguages}.
\newblock In \emph{Proceedings of the Eleventh International Conference on
  Language Resources and Evaluation, {LREC} 2018, Miyazaki, Japan, May 7-12,
  2018}. European Language Resources Association {(ELRA)}.

\bibitem[{Gupta et~al.(2007)Gupta, Nenkova, and
  Jurafsky}]{10.5555/1557769.1557825}
Surabhi Gupta, Ani Nenkova, and Dan Jurafsky. 2007.
\newblock \href {https://www.aclweb.org/anthology/P07-2049} {{M}easuring
  {I}mportance and {Q}uery {R}elevance in {T}opic-focused {M}ulti-document
  {S}ummarization}.
\newblock In \emph{Proceedings of the 45th Annual Meeting of the Association
  for Computational Linguistics Companion Volume Proceedings of the Demo and
  Poster Sessions}, pages 193--196, Prague, Czech Republic. Association for
  Computational Linguistics.

\bibitem[{Haghighi et~al.(2009)Haghighi, Blitzer, DeNero, and
  Klein}]{haghighi-etal-2009-better}
Aria Haghighi, John Blitzer, John DeNero, and Dan Klein. 2009.
\newblock \href {https://www.aclweb.org/anthology/P09-1104} {{B}etter {W}ord
  {A}lignments with {S}upervised {ITG} {M}odels}.
\newblock In \emph{Proceedings of the Joint Conference of the 47th Annual
  Meeting of the {ACL} and the 4th International Joint Conference on Natural
  Language Processing of the {AFNLP}}, pages 923--931, Suntec, Singapore.
  Association for Computational Linguistics.

\bibitem[{Heafield(2011)}]{heafield-2011-kenlm}
Kenneth Heafield. 2011.
\newblock \href {https://www.aclweb.org/anthology/W11-2123} {{K}en{LM}:
  {F}aster and {S}maller {L}anguage {M}odel {Q}ueries}.
\newblock In \emph{Proceedings of the Sixth Workshop on Statistical Machine
  Translation}, pages 187--197, Edinburgh, Scotland. Association for
  Computational Linguistics.

\bibitem[{Inel et~al.(2018)Inel, Haralabopoulos, Li, Van~Gysel, Szl\'{a}vik,
  Simperl, Kanoulas, and Aroyo}]{10.1145/3269206.3271779}
Oana Inel, Giannis Haralabopoulos, Dan Li, Christophe Van~Gysel, Zolt\'{a}n
  Szl\'{a}vik, Elena Simperl, Evangelos Kanoulas, and Lora Aroyo. 2018.
\newblock \href {https://doi.org/10.1145/3269206.3271779} {{S}tudying {T}opical
  {R}elevance with {E}vidence-{B}ased {C}rowdsourcing}.
\newblock In \emph{Proceedings of the 27th ACM International Conference on
  Information and Knowledge Management}, CIKM '18, page 1253–1262, New York,
  NY, USA. Association for Computing Machinery.

\bibitem[{Jain and Wallace(2019)}]{jain2019attention}
Sarthak Jain and Byron~C. Wallace. 2019.
\newblock \href {https://doi.org/10.18653/v1/N19-1357} {{A}ttention is not
  {E}xplanation}.
\newblock In \emph{Proceedings of the 2019 Conference of the North {A}merican
  Chapter of the Association for Computational Linguistics: Human Language
  Technologies, Volume 1 (Long and Short Papers)}, pages 3543--3556,
  Minneapolis, Minnesota. Association for Computational Linguistics.

\bibitem[{Joulin et~al.(2018)Joulin, Bojanowski, Mikolov, J{\'e}gou, and
  Grave}]{joulin-etal-2018-loss}
Armand Joulin, Piotr Bojanowski, Tomas Mikolov, Herv{\'e} J{\'e}gou, and
  Edouard Grave. 2018.
\newblock \href {https://doi.org/10.18653/v1/D18-1330} {{L}oss in
  {T}ranslation: {L}earning {B}ilingual {W}ord {M}apping with a {R}etrieval
  {C}riterion}.
\newblock In \emph{Proceedings of the 2018 Conference on Empirical Methods in
  Natural Language Processing}, pages 2979--2984, Brussels, Belgium.
  Association for Computational Linguistics.

\bibitem[{Junczys-Dowmunt(2018)}]{junczys-dowmunt-2018-dual}
Marcin Junczys-Dowmunt. 2018.
\newblock \href {https://doi.org/10.18653/v1/W18-6478} {{D}ual {C}onditional
  {C}ross-{E}ntropy {F}iltering of {N}oisy {P}arallel {C}orpora}.
\newblock In \emph{Proceedings of the Third Conference on Machine Translation:
  Shared Task Papers}, pages 888--895, Belgium, Brussels. Association for
  Computational Linguistics.

\bibitem[{{Kale} et~al.(2018){Kale}, {Kulkarni}, {Patil}, {Haribhakta},
  {Bhattacharjee}, {Mehta}, {Mithran}, and {Kumar}}]{8554458}
S.~{Kale}, A.~{Kulkarni}, R.~{Patil}, Y.~{Haribhakta}, K.~{Bhattacharjee},
  S.~{Mehta}, S.~{Mithran}, and A.~{Kumar}. 2018.
\newblock \href {https://doi.org/10.1109/ICACCI.2018.8554458} {{O}pen-{D}omain
  {Q}uestion {A}nswering using {F}eature {E}ncoded {D}ynamic {C}oattention
  {N}etworks}.
\newblock In \emph{2018 International Conference on Advances in Computing,
  Communications and Informatics (ICACCI)}, pages 1058--1062.

\bibitem[{Kamholz et~al.(2014)Kamholz, Pool, and Colowick}]{kamholz2014panlex}
David Kamholz, Jonathan Pool, and Susan~M. Colowick. 2014.
\newblock \href
  {http://www.lrec-conf.org/proceedings/lrec2014/summaries/1029.html}
  {{P}anlex: {B}uilding a {R}esource for {P}anlingual {L}exical {T}ranslation}.
\newblock In \emph{Proceedings of the Ninth International Conference on
  Language Resources and Evaluation, {LREC} 2014, Reykjavik, Iceland, May
  26-31, 2014}, pages 3145--3150. European Language Resources Association
  {(ELRA)}.

\bibitem[{Kaszkiel and Zobel(1997)}]{10.1145/278459.258561}
Marcin Kaszkiel and Justin Zobel. 1997.
\newblock \href {https://doi.org/10.1145/258525.258561} {{P}assage {R}etrieval
  {R}evisited}.
\newblock In \emph{Proceedings of the 20th Annual International ACM SIGIR
  Conference on Research and Development in Information Retrieval}, SIGIR '97,
  page 178–185, New York, NY, USA. Association for Computing Machinery.

\bibitem[{Kingma and Ba(2015)}]{kingma2014adam}
Diederik~P. Kingma and Jimmy Ba. 2015.
\newblock \href {http://arxiv.org/abs/1412.6980} {Adam: {A} {M}ethod for
  {S}tochastic {O}ptimization}.
\newblock In \emph{3rd International Conference on Learning Representations,
  {ICLR} 2015, San Diego, CA, USA, May 7-9, 2015, Conference Track
  Proceedings}.

\bibitem[{Koehn et~al.(2007)Koehn, Hoang, Birch, Callison-Burch, Federico,
  Bertoldi, Cowan, Shen, Moran, Zens, Dyer, Bojar, Constantin, and
  Herbst}]{koehn2007moses}
Philipp Koehn, Hieu Hoang, Alexandra Birch, Chris Callison-Burch, Marcello
  Federico, Nicola Bertoldi, Brooke Cowan, Wade Shen, Christine Moran, Richard
  Zens, Chris Dyer, Ond{\v{r}}ej Bojar, Alexandra Constantin, and Evan Herbst.
  2007.
\newblock \href {https://www.aclweb.org/anthology/P07-2045} {{M}oses: {O}pen
  {S}ource {T}oolkit for {S}tatistical {M}achine {T}ranslation}.
\newblock In \emph{Proceedings of the 45th Annual Meeting of the Association
  for Computational Linguistics Companion Volume Proceedings of the Demo and
  Poster Sessions}, pages 177--180, Prague, Czech Republic. Association for
  Computational Linguistics.

\bibitem[{Lample et~al.(2018)Lample, Conneau, Ranzato, Denoyer, and
  J{\'{e}}gou}]{conneau2017word}
Guillaume Lample, Alexis Conneau, Marc'Aurelio Ranzato, Ludovic Denoyer, and
  Herv{\'{e}} J{\'{e}}gou. 2018.
\newblock \href {https://openreview.net/forum?id=H196sainb} {{W}ord
  {T}ranslation without {P}arallel {D}ata}.
\newblock In \emph{6th International Conference on Learning Representations,
  {ICLR} 2018, Vancouver, BC, Canada, April 30 - May 3, 2018, Conference Track
  Proceedings}. OpenReview.net.

\bibitem[{Levy et~al.(2017)Levy, Søgaard, and Goldberg}]{Levy_2017}
Omer Levy, Anders Søgaard, and Yoav Goldberg. 2017.
\newblock \href {https://doi.org/10.18653/v1/e17-1072} {{A} {S}trong {B}aseline
  for {L}earning {C}ross-{L}ingual {W}ord {E}mbeddings from {S}entence
  {A}lignments}.
\newblock \emph{Proceedings of the 15th Conference of the European Chapter of
  the Association for Computational Linguistics: Volume 1, Long Papers}.

\bibitem[{Litschko et~al.(2019)Litschko, Glava\v{s}, Vulic, and
  Dietz}]{10.1145/3331184.3331324}
Robert Litschko, Goran Glava\v{s}, Ivan Vulic, and Laura Dietz. 2019.
\newblock \href {https://doi.org/10.1145/3331184.3331324} {{E}valuating
  {R}esource-{L}ean {C}ross-{L}ingual {E}mbedding {M}odels in {U}nsupervised
  {R}etrieval}.
\newblock In \emph{Proceedings of the 42nd International ACM SIGIR Conference
  on Research and Development in Information Retrieval}, SIGIR’19, page
  1109–1112, New York, NY, USA. Association for Computing Machinery.

\bibitem[{Litschko et~al.(2018)Litschko, Glavaš, Ponzetto, and
  Vulić}]{litschko2018unsupervised}
Robert Litschko, Goran Glavaš, Simone~Paolo Ponzetto, and Ivan Vulić. 2018.
\newblock \href {https://doi.org/10.1145/3209978.3210157} {{U}nsupervised
  {C}ross-{L}ingual {I}nformation {R}etrieval {U}sing {M}onolingual {D}ata
  {O}nly}.
\newblock \emph{The 41st International ACM SIGIR Conference on Research \&
  Development in Information Retrieval}.

\bibitem[{Liu and Croft(2002)}]{10.1145/584792.584854}
Xiaoyong Liu and W.~Bruce Croft. 2002.
\newblock \href {https://doi.org/10.1145/584792.584854} {{P}assage {R}etrieval
  {B}ased on {L}anguage {M}odels}.
\newblock In \emph{Proceedings of the Eleventh International Conference on
  Information and Knowledge Management}, CIKM '02, page 375–382, New York,
  NY, USA. Association for Computing Machinery.

\bibitem[{Loshchilov and Hutter(2019)}]{loshchilov2017decoupled}
Ilya Loshchilov and Frank Hutter. 2019.
\newblock \href {https://openreview.net/forum?id=Bkg6RiCqY7} {{D}ecoupled
  {W}eight {D}ecay {R}egularization}.
\newblock In \emph{7th International Conference on Learning Representations,
  {ICLR} 2019, New Orleans, LA, USA, May 6-9, 2019}. OpenReview.net.

\bibitem[{Luong et~al.(2015)Luong, Pham, and
  Manning}]{luong-etal-2015-bilingual}
Thang Luong, Hieu Pham, and Christopher~D. Manning. 2015.
\newblock \href {https://doi.org/10.3115/v1/W15-1521} {{B}ilingual {W}ord
  {R}epresentations with {M}onolingual {Q}uality in {M}ind}.
\newblock In \emph{Proceedings of the 1st Workshop on Vector Space Modeling for
  Natural Language Processing}, pages 151--159, Denver, Colorado. Association
  for Computational Linguistics.

\bibitem[{Mikolov et~al.(2013)Mikolov, Chen, Corrado, and
  Dean}]{mikolov2013efficient}
Tom{\'{a}}s Mikolov, Kai Chen, Greg Corrado, and Jeffrey Dean. 2013.
\newblock \href {http://arxiv.org/abs/1301.3781} {{E}fficient {E}stimation of
  {W}ord {R}epresentations in {V}ector {S}pace}.
\newblock In \emph{1st International Conference on Learning Representations,
  {ICLR} 2013, Scottsdale, Arizona, USA, May 2-4, 2013, Workshop Track
  Proceedings}.

\bibitem[{Nie(2010)}]{nie2010cross}
Jian-yun Nie. 2010.
\newblock \href {https://doi.org/10.2200/S00266ED1V01Y201005HLT008}
  {{C}ross-{L}anguage {I}nformation {R}etrieval}.
\newblock \emph{Synthesis Lectures on Human Language Technologies}, 3:1--125.

\bibitem[{Niu et~al.(2018)Niu, Denkowski, and Carpuat}]{niu-etal-2018-bi}
Xing Niu, Michael Denkowski, and Marine Carpuat. 2018.
\newblock \href {https://doi.org/10.18653/v1/w18-2710} {{B}i-{D}irectional
  {N}eural {M}achine {T}ranslation with {S}ynthetic {P}arallel {D}ata}.
\newblock \emph{Proceedings of the 2nd Workshop on Neural Machine Translation
  and Generation}.

\bibitem[{Och and Ney(2003)}]{och2003}
Franz~Josef Och and Hermann Ney. 2003.
\newblock \href {https://doi.org/10.1162/089120103321337421} {A {S}ystematic
  {C}omparison of {V}arious {S}tatistical {A}lignment {M}odels}.
\newblock \emph{Computational Linguistics}, 29(1):19--51.

\bibitem[{Radovanovi\'{c} et~al.(2010)Radovanovi\'{c}, Nanopoulos, and
  Ivanovi\'{c}}]{10.1145/1835449.1835482}
Milos Radovanovi\'{c}, Alexandros Nanopoulos, and Mirjana Ivanovi\'{c}. 2010.
\newblock \href {https://doi.org/10.1145/1835449.1835482} {{O}n the {E}xistence
  of {O}bstinate {R}esults in {V}ector {S}pace {M}odels}.
\newblock In \emph{Proceedings of the 33rd International ACM SIGIR Conference
  on Research and Development in Information Retrieval}, SIGIR ’10, page
  186–193, New York, NY, USA. Association for Computing Machinery.

\bibitem[{Ragni and Gales(2018)}]{Ragni2018}
Anton Ragni and Mark Gales. 2018.
\newblock \href {https://doi.org/10.21437/Interspeech.2018-1085} {{A}utomatic
  {S}peech {R}ecognition {S}ystem {D}evelopment in the ``{W}ild"}.
\newblock In \emph{Proc. Interspeech 2018}, pages 2217--2221.

\bibitem[{Ramachandran et~al.(2017)Ramachandran, Liu, and
  Le}]{ramachandran-etal-2017-unsupervised}
Prajit Ramachandran, Peter Liu, and Quoc Le. 2017.
\newblock \href {https://doi.org/10.18653/v1/D17-1039} {{U}nsupervised
  {P}retraining for {S}equence to {S}equence {L}earning}.
\newblock In \emph{Proceedings of the 2017 Conference on Empirical Methods in
  Natural Language Processing}, pages 383--391, Copenhagen, Denmark.
  Association for Computational Linguistics.

\bibitem[{Rubino(2020)}]{rubino-2020-effect}
Carl Rubino. 2020.
\newblock \href {https://www.aclweb.org/anthology/2020.clssts-1.1} {{T}he
  {E}ffect of {L}inguistic {P}arameters in {CLIR} {P}erformance}.
\newblock In \emph{Proceedings of the workshop on Cross-Language Search and
  Summarization of Text and Speech (CLSSTS2020)}, pages 1--6, Marseille,
  France. European Language Resources Association.

\bibitem[{Strohman et~al.(2005)Strohman, Metzler, Turtle, and
  Croft}]{strohmanICIA05}
Trevor Strohman, Donald Metzler, Howard Turtle, and W~Bruce Croft. 2005.
\newblock \href
  {http://citeseerx.ist.psu.edu/viewdoc/download?doi=10.1.1.65.3502&rep=rep1&type=pdf}
  {Indri: {A} {L}anguage {M}odel-based {S}earch {E}ngine for {C}omplex
  {Q}ueries}.
\newblock In \emph{Proceedings of the international conference on intelligent
  analysis}, volume~2, pages 2--6. Citeseer.

\bibitem[{Tiedemann(2012)}]{tiedemann2012parallel}
J{\"{o}}rg Tiedemann. 2012.
\newblock \href
  {http://www.lrec-conf.org/proceedings/lrec2012/summaries/463.html}
  {{P}arallel {D}ata, {T}ools and {I}nterfaces in {OPUS}}.
\newblock In \emph{Proceedings of the Eighth International Conference on
  Language Resources and Evaluation, {LREC} 2012, Istanbul, Turkey, May 23-25,
  2012}, pages 2214--2218. European Language Resources Association {(ELRA)}.

\bibitem[{Vuli\'{c} and Moens(2015)}]{vulic2015monolingual}
Ivan Vuli\'{c} and Marie-Francine Moens. 2015.
\newblock \href {https://doi.org/10.1145/2766462.2767752} {{M}onolingual and
  {C}ross-{L}ingual {I}nformation {R}etrieval {M}odels {B}ased on ({B}ilingual)
  {W}ord {E}mbeddings}.
\newblock In \emph{Proceedings of the 38th International ACM SIGIR Conference
  on Research and Development in Information Retrieval}, SIGIR ’15, page
  363–372, New York, NY, USA. Association for Computing Machinery.

\bibitem[{Wade and Allan(2005)}]{Wade05passageretrieval}
Courtney Wade and James Allan. 2005.
\newblock \href
  {http://citeseerx.ist.psu.edu/viewdoc/summary?doi=10.1.1.152.777} {{P}assage
  {R}etrieval and {E}valuation}.
\newblock Technical report.

\bibitem[{Wolf et~al.(2019)Wolf, Debut, Sanh, Chaumond, Delangue, Moi, Cistac,
  Rault, Louf, Funtowicz, and Brew}]{Wolf2019HuggingFacesTS}
Thomas Wolf, Lysandre Debut, Victor Sanh, Julien Chaumond, Clement Delangue,
  Anthony Moi, Pierric Cistac, Tim Rault, R{\'{e}}mi Louf, Morgan Funtowicz,
  and Jamie Brew. 2019.
\newblock \href {http://arxiv.org/abs/1910.03771} {{H}ugging{F}ace's
  {T}ransformers: {S}tate-of-the-art {N}atural {L}anguage {P}rocessing}.
\newblock \emph{CoRR}, abs/1910.03771.

\bibitem[{Xu and Weischedel(2000)}]{xu2000cross}
Jinxi Xu and Ralph Weischedel. 2000.
\newblock \href {https://doi.org/10.3115/1117794.1117806} {{C}ross-{L}ingual
  {I}nformation {R}etrieval {U}sing {H}idden {M}arkov {M}odels}.
\newblock In \emph{Proceedings of the 2000 Joint SIGDAT Conference on Empirical
  Methods in Natural Language Processing and Very Large Corpora: Held in
  Conjunction with the 38th Annual Meeting of the Association for Computational
  Linguistics - Volume 13}, EMNLP ’00, page 95–103, USA. Association for
  Computational Linguistics.

\bibitem[{Zbib et~al.(2019)Zbib, Zhao, Karakos, Hartmann, DeYoung, Huang,
  Jiang, Rivkin, Zhang, Schwartz, and Makhoul}]{zbib2019neural}
Rabih Zbib, Lingjun Zhao, Damianos Karakos, William Hartmann, Jay DeYoung,
  Zhongqiang Huang, Zhuolin Jiang, Noah Rivkin, Le~Zhang, Richard Schwartz, and
  John Makhoul. 2019.
\newblock \href {https://doi.org/10.1145/3331184.3331222} {{N}eural-{N}etwork
  {L}exical {T}ranslation for {C}ross-{L}ingual {IR} from {T}ext and {S}peech}.
\newblock In \emph{Proceedings of the 42nd International ACM SIGIR Conference
  on Research and Development in Information Retrieval}, SIGIR’19, page
  645–654, New York, NY, USA. Association for Computing Machinery.

\bibitem[{Zhang et~al.(2019)Zhang, Westerfield, Shim, Bingham, Fabbri, Hu,
  Verma, and Radev}]{Zhang_2019}
Rui Zhang, Caitlin Westerfield, Sungrok Shim, Garrett Bingham, Alexander
  Fabbri, William Hu, Neha Verma, and Dragomir Radev. 2019.
\newblock \href {https://doi.org/10.18653/v1/p19-1306} {{I}mproving
  {L}ow-{R}esource {C}ross-lingual {D}ocument {R}etrieval by {R}eranking with
  {D}eep {B}ilingual {R}epresentations}.
\newblock \emph{Proceedings of the 57th Annual Meeting of the Association for
  Computational Linguistics}.

\end{thebibliography}

\clearpage

\appendix
\section{Extra Training Dataset Details}
\label{app:data-details}
When we train SECLR and SECLR-RT via data augmentation, we randomly split the parallel corpus into train set (96\%), validation set (3\%) and test set (1\%). We then use the dataset augmentation technique introduced in Section~\ref{datagen} to generate positive and negative samples for each set. Augmenting the dataset upon the split corpus allows us to achieve more independence between train/validation/test set compared to splitting the dataset augmented on the entire parallel corpus. Note that we only use the validation set for early stopping but we do not tune hyperparameters with the validation set.

We preprocess the parallel corpus, the query collection and the sentence collection with the Moses toolkit \cite{koehn2007moses}. The same preprocessing steps are used for all four languages (English, Somali, Swahili, Tagalog). First, we use Moses puncutation normalizer to normalize the raw text. Second, we use the Moses tokenizer to tokenize the normalized text. Finally, we remove the diacritics in the tokenized text as a cleaning step. 

\section{Examples of Evaluation Data}
\label{app:examples-of-eval}
In this section we demonstrate some examples from the MATERIAL dataset used for evaluation. Example queries include: ``evidence'', ``human rights'', ``chlorine'', ``academy'', ``ratify'', ``constitution'', ``carnage'' and ``Kenya''. On average only 0.13\% of the documents in the Eval collection are relevant to each query, which makes the task hard. 

Here are two examples from Somali Analysis text. Because the documents are long, here we only include the relevant segment of a long relevant document. In the first example, the English query is ``contravention'' and the relevant segment of a long relevant document (translated from Somali to English by human) is ``the security forces captured military equipment coming into the country illegally.'' This segment is relevant to the query because of the word ``illegally''. 

Here is another example where the the English query is ``integrity''. The relevant segment of a long relevant document (translated from Somali to English by human) is ``Hargeisa (Dawan) - Ahmed Mohamed Diriye (Nana) the member of parliament who is part of the Somaliland house of representatives has accused the opposition parties (Waddani and UCID) of engaging in acts of national destruction, that undermines the existence and sovereignty of the country of Somaliland.'' This segment is relevant to the query because of the word ``sovereignty''.

Since there are multiple ways to translate a word and since MT performance is relatively poor in low-resource settings, the task is far more challenging than a simple lexical match between queries and translated documents. 

\section{Extra Experimental Details}
\label{app:supp}
In this section we include extra implementation and experiment details that are not included in the main paper. Information already included in the main paper are not repeated here for conciseness.
\subsection{Model and Training Details}
We train our SECLR and SECLR-RT models on Tesla V100 GPUs. Each model is trained on a single GPU. We report training time of SECLR and SECLR-RT on Somali, Swahili and Tagalog in Table~\ref{tab:training-time}. 

\begin{table}[h]
\centering
\begin{tabular}{lccc}
\toprule
 & Somali & Swahili & Tagalog \\ 
\midrule
SECLR & 77 & 112 & 124  \\ 
SECLR-RT & 179 & 254 & 319  \\ 
\bottomrule
\end{tabular}
\caption{Training time of SECLR and SECLR-RT on Somali, Swahili and Tagalog respectively (in minutes).}
\label{tab:training-time}
\end{table}

As is discussed in Section~\ref{sec:seclrmodel}, the only trainable model parameters of SECLR and SECLR-RT are the word embedding matrices. Thus, SECLR and SECLR-RT have the same number of model parameters. We report the number of trainable parameters of both models on Somali, Swahili and Tagalog in Table~\ref{tab:num-params}. 

\begin{table}[h]
\centering
\begin{tabular}{lccc}
\toprule
 & Somali & Swahili & Tagalog \\ 
\midrule
\# Params. & 14.03M & 22.31M & 21.35M  \\ 
\bottomrule
\end{tabular}
\caption{Number of trainable model parameters of SECLR/SECLR-RT on Somali, Swahili and Tagalog. ``M'' stands for million.}
\label{tab:num-params}
\end{table}

We used Mean Average Precision (MAP) as the evaluation metric in this work. We use the following implementation to compute MAP: \url{https://trec.nist.gov/trec_eval/}.

\subsection{MT Baseline Details}
\label{app:mt-baseline}
\begin{table*}[t]
\begin{center}
\begin{tabular}{ l c c c c c  c  c c  c c  c  c }
\toprule
& \multicolumn{4}{c}{Somali} & \multicolumn{4}{c}{Swahili} & \multicolumn{4}{c}{Tagalog}\\ 
\cmidrule(lr){2-5}\cmidrule(lr){6-9}\cmidrule(lr){10-13} 
 & \multicolumn{2}{c}{Analysis} &  \multicolumn{2}{c}{Dev} & \multicolumn{2}{c}{Analysis} &  \multicolumn{2}{c}{Dev} & \multicolumn{2}{c}{Analysis} &  \multicolumn{2}{c}{Dev} \\  
 \cmidrule(lr){2-3} \cmidrule(lr){4-5} \cmidrule(lr){6-7} \cmidrule(lr){8-9} \cmidrule(lr){10-11} \cmidrule(lr){12-13}
Method & T &  S & T & S & T & S & T &  S & T & S & T & S\\
\midrule
With LSTM & 16.3 & 14.5 & 11.9 & 12.0 & 27.5 & 27.0 & 19.5 & 25.1 & 29.7 & 29.7 & 23.0 & 27.1 \\
No LSTM & \textbf{27.8} & \textbf{24.4} & \textbf{23.0} & \textbf{17.4} & \textbf{43.8} & \textbf{37.9} & \textbf{40.3} & \textbf{38.1} & \textbf{46.7} & \textbf{45.0} & \textbf{49.3} & \textbf{33.9} \\
\hline
\end{tabular}
\caption{Document-level MAP scores for text (T) and speech (S) of the SECLR model with and without LSTM. 
} 
\label{tab:mapmat_lstm}
\end{center}
\end{table*}
\begin{table*}[t]
\begin{center}
\begin{tabular}{ l c c c c c  c  c c  c c  c  c }
\toprule
& \multicolumn{4}{c}{Somali} & \multicolumn{4}{c}{Swahili} & \multicolumn{4}{c}{Tagalog}\\ 
\cmidrule(lr){2-5}\cmidrule(lr){6-9}\cmidrule(lr){10-13} 
 & \multicolumn{2}{c}{Analysis} &  \multicolumn{2}{c}{Dev} & \multicolumn{2}{c}{Analysis} &  \multicolumn{2}{c}{Dev} & \multicolumn{2}{c}{Analysis} &  \multicolumn{2}{c}{Dev} \\  
 \cmidrule(lr){2-3} \cmidrule(lr){4-5} \cmidrule(lr){6-7} \cmidrule(lr){8-9} \cmidrule(lr){10-11} \cmidrule(lr){12-13}
Embed. Init. & T &  S & T & S & T & S & T &  S & T & S & T & S\\
\midrule
Cross-lingual & 35.3 & 27.5 & \textbf{31.1} & \textbf{23.2} & \textbf{48.8} & 41.1 & \textbf{42.5} & 41.6 & 56.3 & 51.1 & 53.8 & 45.3 \\
Monolingual & \textbf{35.4} & \textbf{28.4} & 29.5 & 22.0 & 48.3 & \textbf{48.1} & 39.6 & \textbf{45.4} & \textbf{61.1} & \textbf{55.5} & \textbf{59.0} & \textbf{45.7} \\
\hline
\end{tabular}
\caption{Document-level MAP scores for text (T) and speech (S) of the SECLR-RT model with monolingual or cross-lingual (SID-SGNS) word embedding initialization.
} 
\label{tab:mapmat_sidsgns_init}
\end{center}
\end{table*}
For NMT we train bidirectional MT systems with a 6-layer Transformer architecture with model size of 512, feed-forward network size of 2048, 8 attention heads, and residual connections. We adopt layer normalization and label smoothing. We tie the output weight matrix with the source and target embeddings. We use Adam optimizer with a batch size of 2048 words. We checkpoint models every 1000 updates. Training stops after 20 checkpoints without improvement. During inference, the beam size is set to 5.

Our SMT system uses the following feature functions: phrase translation model, distance-based reordering model, lexicalized reordering model, 5-gram language model on the target side, word penalty, distortion, unknown word penalty and phrase penalty. 

We use backtranslation in earlier versions of MT systems. Following previous work \cite{niu-etal-2018-bi}, we train a bidirectional NMT model that backtranslates source or target monolingual data without an auxiliary model. This backtranslation-based model was the state-of-the-art MT model on Somali and Swahili when the above paper is published. 

Later, we discover that decoder pretraining with monolingual data achieves better performance compared to backtranslation. The decoder pretraining scheme we use now is most similar to the paper by \citet{ramachandran-etal-2017-unsupervised}, where the authors show state-of-the-art results on the WMT English to German translation task with decoder pretraining. 

There is no WMT benchmark for Somali, Swahili or Tagalog, but we use state-of-the-art techniques in our MT systems. We have also experimented with the bilingual data selection method \cite{junczys-dowmunt-2018-dual}. However, this technique does not work well, mostly because low-resource MT systems are not good enough to do scoring.

\section{Extra Experimental Results}
In this section we include extra experimental results that are not included in the main text due to limited space. 

\subsection{SECLR Architecture Exploration}
When we are designing the SECLR model, we experiment with adding LSTMs and using the dot product between LSTM hidden states to compute pairwise similarity between the query and the sentence. We report MAP scores of SECLR with LSTM in Table~\ref{tab:mapmat_lstm}. Experimental results show that adding LSTMs reduces model performance consistently across all three languages. We conjecture that in low-resource settings, contextualized models create spurious correlations (Section~\ref{sec:rationale}). In fact, the XLM-RoBERTa baseline, which captures context effectively via self-attention, also under-performs our SECLR model consistently.

\subsection{Word Embeddings Initialization}
In our SECLR and SECLR-RT models, we initialize word embeddings with monolingual word embeddings in English, Somali, Swahili and Tagalog \cite{mikolov2013efficient, grave2018learning}. One natural question is whether we can achieve performance improvement if we directly initialize with cross-lingual word embeddings. Because SID-SGNS out-performs both Bivec and MUSE consistently by a wide margin (Table~\ref{tab:mapmat_sosw} and Table~\ref{tab:mapmat_tl}), in this experiment we initialize SECLR-RT with the cross-lingual embeddings produced by SID-SGNS. The results of monolingual and cross-lingual embedding initialization (SID-SGNS) are shown in Table~\ref{tab:mapmat_sidsgns_init}. We see that overall monolingual initialization slightly out-performs cross-lingual initialization. Monolingual initialization yields better performance in eight out of 12 Analysis/Dev set conditions and a MAP improvement of 1.7 points when we take the average across Analysis/Dev and all three languages.


\end{document}